\documentclass[11pt,a4paper]{article} 
\usepackage[a4paper,margin=1in]{geometry}

\usepackage[hyphens]{url}  
\usepackage{graphicx} 
\usepackage{natbib} 
\usepackage{caption} 
\usepackage{newfloat}
\usepackage{listings}
\DeclareCaptionStyle{ruled}{labelfont=normalfont,labelsep=colon,strut=off}

\usepackage{booktabs}
\usepackage{threeparttable}
\usepackage{tabularx}

		\title{How Much Do Legal RAG Systems Still Hallucinate?}

\author{
	Souvick Das \and Sallam Abualhaija \and Domenico Bianculli\\
	SnT, University of Luxembourg\\
	Luxembourg\\
	\texttt{\{souvick.das,sallam.abualhaija,domenico.bianculli\}@uni.lu}
}

\begin{document}
			
		\maketitle
			
		\begin{abstract}
		Hallucination is a major challenge for 
retrieval-augmented generation (RAG) systems in 
the legal domain, where ungrounded answers can 
lead to serious consequences. To better understand 
this problem, we conduct a fine-grained analysis 
of hallucination behavior in eight legal RAG 
systems across two legal corpora, the GDPR (in English) 
and a national civil law (in French). Using claim-level and 
answer-level evaluation, 
we report on hallucination density and severity, 
analyze performance across question categories 
and user personas, and validate our findings on an 
independent set of 142 legal-expert-authored questions. 
Our results show that hallucinations remain pervasive, ranging 
from less than 10\% of responses for the best-performing 
systems to nearly half in the worst case. We further find that 
false-premise questions, containing incorrect assumptions 
that must be rejected, produce high 
hallucination rates on the manually-drafted questions.


		\end{abstract}

		\section{Introduction}
\label{sec:introduction}
The growing reliance on generative artificial intelligence (AI) 
systems for knowledge acquisition 
has brought large language models (LLMs) to the forefront of 
modern information retrieval technologies. Despite their 
impressive 
language understanding and generation capabilities, LLMs 
frequently produce \textit{hallucinations}, including factual 
inaccuracies, biased content, and flawed reasoning. Such 
errors raise significant concerns about the reliability and 
trustworthiness of AI-generated information~\cite{orgad2025}. 
Consequently, hallucinations have become a major focus of 
research, attracting substantial attention from the AI 
community~\cite{huang2025,bang2025}.

Hallucinations pose significant risks in the legal domain, where 
inaccurate or ungrounded information can affect 
legal decisions leading to serious financial and societal 
consequences~\cite{dahl2024,magesh2025,banerjee2025,latif2025}.
Several real-world incidents have demonstrated these risks. 
For example, a widely publicized legal case involved 
AI-generated fictitious judicial decisions being cited in a legal 
filing, resulting in judicial sanctions~\cite{gcai2026}. Moreover, 
a recent study reports that even legal research systems 
enhanced with retrieval-augmented generation (RAG) remain 
prone to hallucinations, producing erroneous outputs in 
approximately one out of six queries~\cite{magesh2025}. 
These findings highlight the 
pressing need for reliable methods to detect, evaluate, and 
mitigate hallucinations in legal AI systems.

Addressing hallucinations in the legal domain is particularly 
challenging. Legal reasoning often relies on large collections 
of heterogeneous legal sources, and determining the correct 
legal interpretation is often non-trivial~\cite{mik2024}. As 
a result, hallucinations in legal AI systems cannot always be 
assessed using general-purpose fact-checking criteria, 
motivating 
the need for domain-specific approaches to their analysis and 
mitigation~\cite{hou2024,magesh2025}. Furthermore, 
hallucination behavior can vary substantially depending on the 
type of legal question. Prior studies have shown that LLMs 
often fail to correct users' incorrect legal assumptions in 
counterfactual question settings, suggesting that certain 
question categories are inherently more prone to 
hallucinations~\cite{dahl2024}. Existing work also suggests 
that users with different backgrounds and goals interact 
with AI systems in different ways~\cite{lin2025}. 
Together, these observations motivate a systematic 
investigation of hallucination behavior across question 
categories and user personas.

The widespread adoption of RAG further complicates 
hallucination analysis. Although RAG is designed to reduce 
hallucinations by grounding responses in retrieved evidence, 
a 
hallucinated answer may originate from the retrieval 
component, the generation component, or their 
interaction~\cite{ru2024,das2026}. Consequently, 
evaluating 
and diagnosing hallucinations in legal RAG systems is 
significantly more challenging than assessing standalone 
LLMs.

Recent benchmarks have highlighted the prevalence of 
hallucinations in legal AI systems, including hallucinated 
citations~\cite{liu2026} and responses that are unfaithful to 
retrieved evidence~\cite{yeh2026,hu2026}.
To obtain a more 
diagnostic understanding of these failures, evaluation must 
move beyond aggregate performance scores toward  
\emph{claim-level} analysis, where a claim is a minimal 
factual statement that can be independently verified against 
supporting evidence~\cite{min2023,ru2024}. Recent 
work has adopted this perspective for RAG evaluation, 
factuality assessment, and hallucination detection by 
extracting claims from generated answers and assessing their 
support in the retrieved 
context~\cite{ru2024,hu2024,scire2024}. Such fine-grained 
analysis enables a more detailed characterization of 
hallucination behavior. 

More recently, ClaimRAG-LAW~\cite{das2026} introduced a 
multilingual, multi-jurisdiction benchmark for evaluating legal 
RAG systems through claim-level assessment. While the 
benchmark enables systematic comparison of state-of-the-art 
systems, it provides limited insight into how hallucination 
behavior varies across question categories and user personas. 
Moreover, it remains unclear whether the 
observed hallucination patterns generalize beyond the 
benchmark itself. Consequently, our understanding of when 
and under which conditions legal RAG systems hallucinate 
remains incomplete.

In this work, we address these gaps through a comprehensive 
analysis of hallucination behavior in legal RAG systems. We 
examine hallucinations at both the \emph{answer level} and 
the \emph{claim level}, analyze their variation across question 
categories and user personas, and assess the generalizability 
of the observed patterns through external validation on 
manually-drafted legal questions. Answer-level analysis 
measures how often hallucinations occur in generated 
responses, whereas claim-level analysis captures the density 
and severity of hallucinated claims. This distinction is 
particularly important in legal settings, where a response 
containing a single unsupported claim differs substantially 
from one containing multiple unsupported or contradicted 
claims. By combining these complementary perspectives, we 
provide a more nuanced characterization of the prevalence, 
severity, and distribution of hallucinations in legal RAG 
systems.

This paper makes the following contributions:
\begin{itemize}
	\item We present a comprehensive analysis of hallucination 
	behavior in legal RAG systems. In particular, we evaluate 
	eight state-of-the-art RAG systems on two legal 
	sources provided by ClaimRAG-LAW~\cite{das2026}: the 
	General Data Protection Regulation (GDPR)~\cite{EUGDPR} 
	in English and a national civil law corpus (CIVIL) in 
	French~\cite{CIVIL}.
        The evaluated systems combine two 
	retrievers, BM25~\cite{robertson2009} and 
	E5~\cite{wang2024},
        with four generators: 
	GPT-4~\cite{gpt4}, GPT-5~\cite{singh2025}, 
	Llama-3.1-8B-Instruct (hereafter 
	Llama3-8B)~\cite{grattafiori2024}, and 
	Mixtral-8x7B~\cite{jiang2024}. 
	Our results show that 
	hallucinations remain pervasive across all systems, 
	ranging from 8\% to 20\% of responses for BM25+GPT-5 to 
	nearly half of all responses for Llama3-8B-based systems.
        We further show that most hallucinations 
	are isolated single-claim errors, while severe multi-claim 
	hallucinations occur less frequently but remain present even 
	in top-performing systems.
	
	\item We show that hallucination behavior in legal RAG 
	systems is highly sensitive to both question category and 
	user persona. Across GDPR and CIVIL, BM25+GPT-5 
	consistently achieves the lowest hallucination rates, while 
	Llama3-8B-based systems exhibit the highest 
	rates. We further find that 
	\textit{false-premise} questions (queries containing 
	incorrect assumptions) are
	particularly prone to hallucinations. Moreover, 
	hallucination rates vary across user personas, with questions 
	associated with legal experts generally resulting in fewer 
	hallucinations than those associated with civil officers or citizens, 
	suggesting that query formulation, user 
	intent, and domain knowledge play an important role in 
	influencing the hallucination behavior of legal RAG systems.
	
	\item We validate our 
	findings on an independent set of 142 manually-drafted 
	legal 
	questions spanning GDPR and CIVIL. The resulting trends 
    are similar to the benchmark results, confirming 
	that the observed hallucination patterns generalize beyond 
	ClaimRAG-LAW and are not benchmark-specific. 
\end{itemize}

The remainder of the paper is organized as follows. 
Section~\ref{sec:state_of_the_art} reviews the related work on 
hallucination and legal RAG evaluation. 
Section~\ref{sec:evaluation} describes the empirical setup and 
presents the hallucination analysis across systems. 
Section~\ref{sec:discussion} presents an external validation 
using manually drafted legal questions. 
Section~\ref{sec:threats} discusses the threats to validity 
considerations and limitations of our study. Finally, 
Section~\ref{sec:conclusion} 
concludes the 
paper.


		\section{Related 
	Work}
\label{sec:state_of_the_art}

Hallucination detection has attracted considerable attention 
from the research community, with existing approaches 
broadly categorized as \textit{reference-free} and 
\textit{reference-based} methods. Reference-free approaches 
assess outputs without relying on external knowledge sources, 
often leveraging self-consistency 
measures~\cite{Li2024}. For example, 
SelfCheckGPT~\cite{manakul2023} detects hallucinations by 
measuring consistency across multiple outputs generated by 
the same model. Similar ideas have recently been applied in 
the legal domain, where LLMs are used as factual-consistency 
judges for legal question answering systems, demonstrating 
promising hallucination detection 
capabilities~\cite{hu2025,enguehard2025}.

In contrast, reference-based approaches detect hallucinations 
by comparing model outputs against a trusted source of 
evidence, such as gold-standard answers, retrieved 
documents, or supporting context. As our work falls within this 
category, we focus on reference-based techniques in the 
remainder of this section. 

A prominent class of reference-based methods detects 
hallucinations through claim-level verification. Such methods 
first decompose generated responses into verifiable claims 
and then assess each claim against supporting 
evidence~\cite{min2023,ru2024,galimzianova2025}. 
One of the early examples is FActScore~\cite{min2023}, which 
measures factual accuracy by verifying individual claims 
against retrieved evidence. Building on this idea, 
FENICE~\cite{scire2024} combines claim extraction and 
entailment analysis to assess factual consistency at multiple 
levels of granularity while identifying supporting evidence for 
each claim. RefChecker~\cite{hu2024} further structures 
claims as subject-predicate-object triples and evaluates 
their entailment against a reference context, making it 
particularly suitable for RAG systems. Collectively, these 
methods establish claim-level verification as an effective 
framework 
for fine-grained hallucination analysis. 

Beyond general-purpose hallucination detection, a growing 
body of research has focused on evaluating LLMs in the legal 
domain. Early benchmarks, such as 
LexGLUE~\cite{chalkidis2022}, LEXTREME~\cite{niklaus2023}, 
and LegalBench-RAG~\cite{pipitone2024}, assess legal 
language understanding, reasoning, and retrieval capabilities 
across a range of tasks~\cite{louis2024}. More recently, 
attention has shifted toward the factual reliability of legal RAG 
systems. Representative examples include  
LLeQA~\cite{louis2024}, which studies retrieve-then-read 
question answering over Belgian statutory law, and the work of 
\citet{liu2026}, which evaluates legal hallucinations through 
citation verification in real-world legal filings. However, 
existing benchmarks rarely examine the generalizability of 
their findings 
beyond the benchmark. 

ClaimRAG-LAW~\cite{das2026} is a recent benchmark that 
provides a fine-grained 
framework for evaluating hallucinations in legal RAG systems 
across multiple jurisdictions, languages, question categories, 
and user personas. 
While it enables systematic comparison of legal RAG systems, 
its analysis primarily focuses on aggregate 
hallucination metrics, offering limited insight into how 
hallucination behavior varies across question categories and 
user 
perspectives. Additionally, it does not examine whether the 
observed hallucination patterns persist beyond the benchmark 
questions. Building on ClaimRAG-LAW, we present a 
systematic analysis of hallucination behavior across 
eight state-of-the-art RAG systems, considering different 
question categories, and personas, and further validate our 
findings on an independent set of legal-expert-authored 
questions.


		\section{Empirical Evaluation}
\label{sec:evaluation}

This paper addresses 
the following research questions (\textbf{RQs}):

\paragraph{RQ1. What is the hallucination rate of RAG 
systems in the legal domain? }
This RQ provides an in-depth analysis of the hallucination 
behavior for legal RAG systems on GDPR 
and CIVIL. Specifically, we investigate hallucination 
both at the 
claim-level as well as answer-level to better understand 
whether hallucinations are distributed across responses 
or concentrated in a subset thereof.

\paragraph{RQ2. How do hallucination  rates vary across 
	question categories and user personas?}
This RQ studies the impact of question type and  user
 persona on the hallucination rates of legal RAG systems.


\subsection{Dataset}
\label{subsec:dataset}

We conduct our analysis on 
ClaimRAG-LAW~\cite{das2026},
a multilingual benchmark for fine-grained evaluation of legal 
RAG systems. It is composed of two sub-datasets, collectively 
containing 317 
expert-validated question-answer (QA) pairs 
derived from two legal sources: the \textit{General Data 
Protection Regulation} (GDPR, English; 149 QA pairs) and a 
national Civil Code (CIVIL, French; 168 QA pairs).

The benchmark covers four question categories: 
\textit{General Legal Research} (GR, 268 QA pairs, 84.5\%), 
\textit{Factual 
Recall} (FR, 33 QA pairs, 10.4\%), \textit{False Premise} (FP, 13 
QA pairs, 
4.1\%), and \textit{Jurisdiction/Time-Specific} (JT, 3 QA pairs, 
0.9\%). 
It also 
considers three user personas with different expertise levels: 
\textit{Legal Expert} (LE, 156 QA pairs, 49.2\%), \textit{Civil 
Officer} (CO, 106 QA 
pairs, 33.4\%), and \textit{Citizen} (CI, 55 QA pairs, 17.4\%).

\subsection{Evaluated RAG Systems}
\label{subsec:rag-systems}
We evaluate the same eight RAG systems as 
ClaimRAG-LAW~\cite{das2026}, obtained by combining two 
retrievers (BM25 and E5) with four generators (Llama3-8B, 
Mixtral-8x7B, GPT-4, and GPT-5).

\subsection{Evaluation Procedure} 
\label{subsec:eval-procedure]}
To answer our RQs, we analyze the generated answers of the 
 RAG systems on GDPR and CIVIL datasets,
following the same evaluation setup as 
ClaimRAG-LAW~\cite{das2026}. 
For each question, we examine the generated answer, 
corresponding 
ground-truth answer, extracted claims, 
entailment labels, question category, and persona without 
re-executing the underlying RAG systems. 

We report \textit{claim-level hallucination rate (CL)}, defined 
as the proportion of hallucinated claims among all generated 
claims, and \textit{answer-level hallucination rate (AL)}, 
defined as the proportion of answers containing at least one 
hallucinated claim (i.e., a neutral or contradictory 
claim)~\cite{ru2024}. Thus, CL measures 
hallucination density, whereas AL measures hallucination 
frequency.

We additionally report \textit{strict claim-level} (CL$^s$) and 
\textit{strict 	answer-level} (AL$^s$) hallucination rates, which 
count only 
explicitly contradicted claims as hallucinations~\cite{hu2024}. 
These stricter metrics provide a conservative estimate of 
hallucination behavior by focusing on factually incorrect legal 
claims rather than merely unsupported content.

Finally, beyond the above metrics, we analyze the distribution 
of hallucinated claims within hallucinated answers to 
determine whether hallucinations are concentrated in a small 
number of severely affected answers or spread more evenly 
across a system's responses.

The evaluation materials and source code will be released 
upon publication under an appropriate open-source license.


		\subsection{Legal 
	RAG 
	Hallucination Rates (RQ1)}
\label{sec:rq1}

\begin{table*}[t]
	\centering
	\footnotesize
	\setlength{\tabcolsep}{4pt}
	
	\begin{tabular*}{\textwidth}{@{\extracolsep{\fill}}lcccccccccc}
		\toprule
		
		& \multicolumn{5}{c}{\textbf{GDPR}}
		& \multicolumn{5}{c}{\textbf{CIVIL}} \\
		
		\cmidrule(lr){2-6}
		\cmidrule(lr){7-11}
		
		\textbf{RAG System}
		& $M$
		& CL$\downarrow$
		& CL$^{s}\downarrow$
		& AL$\downarrow$
		& AL$^{s}\downarrow$
		& $M$
		& CL$\downarrow$
		& CL$^{s}\downarrow$
		& AL$\downarrow$
		& AL$^{s}\downarrow$ \\
		
		\midrule
		
		BM25+Llama3-8B
		& $10.6{\scriptstyle\pm5.6}$
		& $4.6{\scriptstyle\pm8.6}$
		& $3.5{\scriptstyle\pm8.5}$
		& $29.5$
		& $20.8$
		& $6.1{\scriptstyle\pm2.6}$
		& $17.9{\scriptstyle\pm25.8}$
		& $5.0{\scriptstyle\pm10.1}$
		& $45.8$
		& $24.4$ \\
		
		BM25+Mixtral-8x7B
		& $10.1{\scriptstyle\pm5.8}$
		& $3.4{\scriptstyle\pm9.3}$
		& $3.3{\scriptstyle\pm7.5}$
		& $\mathbf{20.1}$
		& $20.8$
		& $6.4{\scriptstyle\pm3.2}$
		& $16.4{\scriptstyle\pm24.4}$
		& $6.6{\scriptstyle\pm14.1}$
		& $41.1$
		& $27.9$ \\
		
		BM25+GPT-4
		& $13.5{\scriptstyle\pm6.7}$
		& $2.9{\scriptstyle\pm5.0}$
		& $2.3{\scriptstyle\pm7.3}$
		& $30.9$
		& $14.7$
		& $6.9{\scriptstyle\pm3.6}$
		& $4.1{\scriptstyle\pm14.6}$
		& $\mathbf{4.4}{\scriptstyle\pm11.0}$
		& $15.5$
		& $20.8$ \\
		
		BM25+GPT-5
		& $17.9{\scriptstyle\pm12.2}$
		& $\mathbf{1.5{\scriptstyle\pm3.8}}$
		& $\mathbf{0.9{\scriptstyle\pm4.6}}$
		& $\mathbf{20.1}$
		& $10.7$
		& $7.5{\scriptstyle\pm5.6}$
		& $\mathbf{1.2{\scriptstyle\pm4.7}}$
		& $6.1{\scriptstyle\pm18.2}$
		& $\mathbf{8.3}$
		& $16.6$ \\
		
		E5+Llama3-8B
		& $9.2{\scriptstyle\pm4.4}$
		& $9.5{\scriptstyle\pm12.4}$
		& $3.4{\scriptstyle\pm6.8}$
		& $48.3$
		& $24.1$
		& $5.7{\scriptstyle\pm3.0}$
		& $11.2{\scriptstyle\pm21.1}$
		& $8.8{\scriptstyle\pm18.7}$
		& $32.1$
		& $27.9$ \\
		
		E5+Mixtral-8x7B
		& $6.0{\scriptstyle\pm3.0}$
		& $4.7{\scriptstyle\pm10.1}$
		& $2.6{\scriptstyle\pm8.3}$
		& $21.5$
		& $10.7$
		& $4.0{\scriptstyle\pm1.8}$
		& $4.7{\scriptstyle\pm15.3}$
		& $6.8{\scriptstyle\pm16.1}$
		& $13.7$
		& $20.2$ \\
		
		E5+GPT-4
		& $6.9{\scriptstyle\pm3.1}$
		& $3.0{\scriptstyle\pm7.0}$
		& $2.5{\scriptstyle\pm8.0}$
		& $\mathbf{20.1}$
		& $11.4$
		& $4.3{\scriptstyle\pm1.9}$
		& $3.3{\scriptstyle\pm12.1}$
		& $\mathbf{4.4}{\scriptstyle\pm13.0}$
		& $9.5$
		& $\mathbf{14.8}$ \\
		
		E5+GPT-5
		& $9.5{\scriptstyle\pm6.4}$
		& $5.6{\scriptstyle\pm9.5}$
		& $1.5{\scriptstyle\pm7.8}$
		& $36.2$
		& $\mathbf{5.3}$
		& $6.2{\scriptstyle\pm4.1}$
		& $4.9{\scriptstyle\pm12.7}$
		& $6.3{\scriptstyle\pm16.9}$
		& $22.0$
		& $20.8$ \\
		
		\bottomrule
	\end{tabular*}
	
	\caption{Hallucination rates (\%) of RAG systems in legal 
		application context (\textbf{RQ1}).
		$M$ denotes the average number of claims per 
		response
		(mean $\pm$ standard deviation). 
		$\downarrow$ indicates that 
		lower values
		are better.}
	\label{tab:hallucination_results}
	
\end{table*}

\begin{table*}[t]
	\centering
	\footnotesize
	\setlength{\tabcolsep}{4.75pt}
	
	\begin{tabular}{l rrrrrrrrrr rrrrrrrrrr}
		\toprule
		
		& \multicolumn{10}{c}{\textbf{CIVIL}}
		& \multicolumn{10}{c}{\textbf{GDPR}} \\
		
		\cmidrule(lr){2-11}
		\cmidrule(lr){12-21}
		
		\textbf{RAG System}
		& H0 & H1 & H2 & H3 & H4 & H5 & H6 & H7 
		& 
		H8 
		& 
		H9
		& H0 & H1 & H2 & H3 & H4 & H5 & H6 & H7 
		& 
		H8 
		& 
		H9 \\
		
		\midrule
		
		BM25+Llama3-8B
		& 91 & 33 & 23 & 13 & 4 & 4 & 0 & 0 & 0 & 0
		& 105 & 27 & 13 & 3 & 0 & 1 & 0 & 0 & 0 & 0 
		\\
		
		BM25+Mixtral-8x7B
		& 99 & 30 & 12 & 19 & 3 & 2 & 2 & 0 & 1 & 0
		& 119 & 21 & 7 & 1 & 0 & 1 & 0 & 0 & 0 & 0 \\
		
		BM25+GPT-4
		& 142 & 23 & 0 & 3 & 0 & 0 & 0 & 0 & 0 & 0
		& 103 & 32 & 11 & 3 & 0 & 0 & 0 & 0 & 0 & 0 
		\\
		
		BM25+GPT-5
		& 154 & 11 & 1 & 1 & 1 & 0 & 0 & 0 & 0 & 0
		& 119 & 15 & 11 & 3 & 0 & 1 & 0 & 0 & 0 & 0 \\
		
		E5+Llama3-8B
		& 114 & 29 & 19 & 4 & 1 & 0 & 1 & 0 & 0 & 0
		& 77 & 38 & 20 & 9 & 3 & 2 & 0 & 0 & 0 & 0 
		\\
		
		E5+Mixtral-8x7B
		& 145 & 18 & 3 & 2 & 0 & 0 & 0 & 0 & 0 & 0
		& 117 & 28 & 2 & 2 & 0 & 0 & 0 & 0 & 0 & 0 \\
		
		E5+GPT-4
		& 152 & 10 & 6 & 0 & 0 & 0 & 0 & 0 & 0 & 0
		& 119 & 27 & 3 & 0 & 0 & 0 & 0 & 0 & 0 & 0 \\
		
		E5+GPT-5
		& 131 & 23 & 7 & 3 & 1 & 0 & 2 & 0 & 0 & 1
		& 95 & 32 & 13 & 3 & 3 & 2 & 0 & 0 & 1 & 0 \\
		
		\bottomrule
	\end{tabular}
	
	\caption{Distribution of hallucinated claims per 
		answer (\textbf{RQ1}). }
	\label{tab:hc_distribution}
\end{table*}

Table~\ref{tab:hallucination_results} reports the hallucination 
rates for the evaluated RAG systems according to the metrics 
defined above.
We report the standard deviation (SD) 
values for CL and CL$^s$ but not for AL and AL$^s$, as the 
answer-level metrics are essentially binary indicators of 
whether an answer is marked as hallucinated. We also report 
the average number of claims per response (column $M$) to 
provide context for the claim-level results.

\noindent\textbf{Hallucination Density.}
The table shows that GPT-based RAG 
systems consistently exhibit the lowest CL values, with 
BM25+GPT-5 achieving the best results on both GDPR 
(\(1.5\% 
\pm 
3.8\)) and CIVIL (\(1.2\% \pm 4.7\)). In contrast, 
Llama3-8B-based systems generally produce the highest 
claim-level hallucination rates, with 
E5+Llama3-8B on GDPR (\(9.5\% \pm 12.4\)) and with 
BM25+Llama3-8B on CIVIL (\(17.9\% 
\pm 25.8\)). These results show that hallucination can be 
highly impacted by the retrieval and generator  
components in a RAG system.
Furthermore, the relatively large SD values suggest that 
hallucinations may be concentrated in a subset of the answers 
rather than occurring uniformly across outputs, as discussed 
later in this section.

The average number of claims per generated answer ($M$) 
also varies across RAG systems.  
$M$ provides an indication of response verbosity. For 
example, BM25+GPT-5 
produces the highest average number of claims overall (GDPR: 
\(17.9 \pm 12.2\), CIVIL: \(7.5 \pm 5.6\)). 
In contrast, E5+Llama3-8B generates substantially fewer 
claims compared to GPT-based systems. Despite their greater 
verbosity, GPT-based systems maintain lower hallucination 
rates, suggesting that the higher hallucination rates of 
open-weight models cannot be attributed solely to response 
length.

Differences in verbosity also appear to depend on the retrieval 
component. Across both datasets, E5-based RAG systems 
consistently generate fewer claims per 
answer than BM25-based ones, highlighting that verbosity 
could partially depend on the underlying 
retrieval. The contexts retrieved by E5 appear to encourage 
generators to produce more conservative and concise 
answers, unlike BM25 retrieval, which generally encourages 
more elaborate responses.

Furthermore, the gap between CL and CL$^s$ values reveals 
distinct hallucination patterns. A large gap, as observed for 
E5+GPT-5 on GDPR (\(5.6\%\) vs. \(1.5\%\)),
suggests that 
most hallucinated claims are rather unsupported rather than 
explicitly contradicted by the evidence.  In 
contrast, a smaller gap, as observed for BM25+Llama3-8B on 
GDPR (\(4.6\%\) vs. \(3.5\%\)), 
indicates a larger proportion of 
explicit contradicted claims. Since contradicted claims 
correspond to factually incorrect legal information, they 
represent a more severe hallucination than simply neutral 
claims 
and would therefore require careful verification, 
ideally by legal experts.

\noindent\textbf{Hallucination Frequency.} The answer-level 
results reveal that hallucinations remain common across all 
evaluated systems. 
BM25+GPT-5 is the most robust against 
hallucinations, 
achieving the lowest AL hallucination rate on GDPR 
($20.1$\%) 
and CIVIL ($8.33$\%). In contrast,  E5+Llama3-8B performs 
worst on GDPR ($48.3$\%), while BM25+Llama3-8B 
records the highest hallucination rate on CIVIL ($45.8$\%). 
These results indicate that, in 
the worst case, BM25+GPT-5 hallucinates in roughly one out 
of five answers, whereas Llama3-8B-based systems 
hallucinate in nearly half of the generated answers.

Taken together, CL and AL values provide a complementary 
view of hallucination behavior. High values for both metrics 
indicate 
frequent and dense hallucinations (many unsupported claims 
in many answers), as observed for E5+Llama3-8B on GDPR 
(AL = $48.3$\%, CL = $9.5$\%\(\pm 12.4\)) and 
BM25+Llama3-8B on CIVIL (AL = $45.8$\%, CL = 
$17.9$\%\(\pm 25.8\)). In contrast, low AL with relatively high 
CL values indicate that 
hallucinations are concentrated in a smaller number of 
responses, e.g., 
E5+GPT-5 on CIVIL (AL = $22.0$\%, CL = $4.9$\%, with a 
standard deviation of \(\pm 12.7\), indicating severe outliers).
conversely, high AL combined with low CL values suggest 
hallucinations 
are spread across many answers, but remain limited in each 
one (i.e., only few unsupported 
claims), e.g., BM25+GPT-4 on GDPR (AL = $30.8$\%, CL = 
$2.9$\%).  This latter pattern is more 
common across the evaluated systems, suggesting that legal 
hallucinations often arise from 
isolated unsupported claims rather than entirely 
hallucinated answers.
The most desirable behavior is characterized by low values for 
both metrics, reflecting infrequent and limited hallucinations.

\paragraph{Hallucination Severity.} To better characterize 
hallucination severity,  
Table~\ref{tab:hc_distribution} reports the 
distribution of unsupported claims per answer (H).
Specifically, 
column H0 denotes 
hallucination-free answers, and columns H$k$ ($k \geq 1$) 
denotes 
answers 
containing exactly $k$ unsupported claims.

Across both datasets, H1 is the dominant category, 
indicating that most hallucinated answers contain only a single 
unsupported claim and are therefore localized rather than 
pervasive errors.
Consistent with the results above, BM25+GPT-5 ranks among 
the most reliable systems, 
achieving the highest number of hallucination-free responses 
on CIVIL (H0 = 154) and tying for the highest count on GDPR 
(H0 = 119), alongside BM25+Mixtral-8x7B and E5+GPT-4.  In 
contrast, E5+Llama3-8B records the lowest H0 
count on GDPR (77), while BM25+Llama3-8B performs worst 
on CIVIL (91). 

Severe hallucinations involving multiple unsupported claims 
remain present even by the 
strongest systems, albeit less frequently. 
For example, E5+GPT-5 occasionally produces answers 
containing eight or more unsupported claims (H8 on GDPR and 
H9 on CIVIL in one answer each). 
Although rare, such cases highlight the potential risk of 
high-severity hallucinations in legal applications and motivate 
fine-grained evaluation beyond aggregate 
metrics. 

Overall, the results suggest that hallucinations in legal RAG 
systems are more often isolated than pervasive, but severe 
multi-claim hallucinations remain a persistent concern.


			\subsection{Hallucination Rate Across 
	Question Categories and Personas (RQ2)}

\begin{table*}[t]
	\centering
	\footnotesize
	\setlength{\tabcolsep}{3pt}
	
	\begin{tabular*}{\textwidth}{@{\extracolsep{\fill}}l
			c@{\hspace{1pt}}c
			c@{\hspace{1pt}}c
			c@{\hspace{1pt}}c
			c@{\hspace{1pt}}c
			c@{\hspace{1pt}}c
			c@{\hspace{1pt}}c}
		\toprule
		
		& \multicolumn{6}{c}{\textbf{GDPR}}
		& \multicolumn{6}{c}{\textbf{CIVIL}} \\
		
		\cmidrule(lr){2-7}
		\cmidrule(lr){8-13}
		
		& \multicolumn{2}{c}{\textbf{GR} ($Q=131$)}
		& \multicolumn{2}{c}{\textbf{FR} ($Q=9$)}
		& \multicolumn{2}{c}{\textbf{FP} ($Q=8$)}
		& \multicolumn{2}{c}{\textbf{GR} ($Q=137$)}
		& \multicolumn{2}{c}{\textbf{FR} ($Q=24$)}
		& \multicolumn{2}{c}{\textbf{FP} ($Q=5$)} \\
		
		\cmidrule(lr){2-3}
		\cmidrule(lr){4-5}
		\cmidrule(lr){6-7}
		\cmidrule(lr){8-9}
		\cmidrule(lr){10-11}
		\cmidrule(lr){12-13}
		
		\textbf{RAG System}
		& CL$\downarrow$ & AL$\downarrow$
		& CL$\downarrow$ & AL$\downarrow$
		& CL$\downarrow$ & AL$\downarrow$
		& CL$\downarrow$ & AL$\downarrow$
		& CL$\downarrow$ & AL$\downarrow$
		& CL$\downarrow$ & AL$\downarrow$ \\
		
		\midrule
		
		BM25+Llama3-8B
		& $4.5{\scriptstyle\pm8.3}$ & $29.8$
		& $10.9{\scriptstyle\pm13.7}$ & $44.4$
		& $0.9{\scriptstyle\pm2.5}$ & $12.5$
		& $13.4{\scriptstyle\pm22.0}$ & $39.4$
		& $35.1{\scriptstyle\pm29.9}$ & $75.0$
		& $56.0{\scriptstyle\pm37.8}$ & $80.0$ \\
		
		BM25+Mixtral-8x7B
		& $3.6{\scriptstyle\pm9.3}$ & $22.1$
		& $\mathbf{0.0{\scriptstyle\pm0.0}}$ & $\mathbf{0.0}$
		& $5.0{\scriptstyle\pm14.1}$ & $12.5$
		& $11.6{\scriptstyle\pm21.1}$ & $32.1$
		& $39.1{\scriptstyle\pm29.4}$ & $75.0$
		& $24.2{\scriptstyle\pm12.4}$ & $100.0$ \\
		
		BM25+GPT-4
		& $2.9{\scriptstyle\pm5.0}$ & $31.3$
		& $1.8{\scriptstyle\pm3.6}$ & $22.2$
		& $3.0{\scriptstyle\pm6.1}$ & $25.0$
		& $4.4{\scriptstyle\pm15.7}$ & $16.1$
		& $2.4{\scriptstyle\pm8.3}$ & $8.3$
		& $6.5{\scriptstyle\pm9.3}$ & $40.0$ \\
		
		BM25+GPT-5
		& $\mathbf{1.6{\scriptstyle\pm3.6}}$ & $22.1$
		& $2.5{\scriptstyle\pm7.4}$ & $11.1$
		& $\mathbf{0.0{\scriptstyle\pm0.0}}$ & $\mathbf{0.0}$
		& $\mathbf{1.4{\scriptstyle\pm5.2}}$ & $\mathbf{10.2}$
		& $\mathbf{0.0{\scriptstyle\pm0.0}}$ & $\mathbf{0.0}$
		& $\mathbf{0.0{\scriptstyle\pm0.0}}$ & $\mathbf{0.0}$ \\
		
		E5+Llama3-8B
		& $9.3{\scriptstyle\pm12.3}$ & $46.6$
		& $13.9{\scriptstyle\pm16.2}$ & $66.7$
		& $9.4{\scriptstyle\pm9.3}$ & $62.5$
		& $12.5{\scriptstyle\pm22.7}$ & $32.9$
		& $4.2{\scriptstyle\pm8.2}$ & $25.0$
		& $13.3{\scriptstyle\pm13.9}$ & $60.0$ \\
		
		E5+Mixtral-8x7B
		& $4.3{\scriptstyle\pm9.4}$ & $\mathbf{20.6}$
		& $13.0{\scriptstyle\pm18.2}$ & $44.4$
		& $1.8{\scriptstyle\pm5.1}$ & $12.5$
		& $5.6{\scriptstyle\pm16.8}$ & $15.3$
		& $\mathbf{0.0{\scriptstyle\pm0.0}}$ & $\mathbf{0.0}$
		& $6.9{\scriptstyle\pm9.6}$ & $40.0$ \\
		
		E5+GPT-4
		& $3.0{\scriptstyle\pm7.1}$ & $\mathbf{20.6}$
		& $1.9{\scriptstyle\pm5.6}$ & $11.1$
		& $4.4{\scriptstyle\pm9.0}$ & $25.0$
		& $3.8{\scriptstyle\pm13.0}$ & $11.0$
		& $\mathbf{0.0{\scriptstyle\pm0.0}}$ & $\mathbf{0.0}$
		& $6.7{\scriptstyle\pm14.9}$ & $20.0$ \\
		
		E5+GPT-5
		& $5.9{\scriptstyle\pm9.1}$ & $39.7$
		& $5.6{\scriptstyle\pm16.7}$ & $11.1$
		& $2.1{\scriptstyle\pm5.9}$ & $12.5$
		& $5.6{\scriptstyle\pm13.3}$ & $25.6$
		& $2.1{\scriptstyle\pm10.2}$ & $4.2$
		& $2.5{\scriptstyle\pm5.6}$ & $20.0$ \\
		
		\bottomrule
	\end{tabular*}
	
	\caption{
		Hallucination rates (\%) across \textit{question 
		categories} (\textbf{RQ2}).
		GR = General Research, FR = Factual Recall, and FP = 
		False Premise.
		CL and AL denote claim-level and answer-level 
		hallucination rates, where $\pm$ 
		is the standard deviation.
	}
	\label{tab:hallucination_per_cat}
	
\end{table*}


\noindent\textbf{Hallucinations Analysis by Question
	Category. }
Table~\ref{tab:hallucination_per_cat} reports CL and AL
across question categories (GR: General Research, FR: Factual 
Recall, 
and FP: False Premise). FR and FP results should be
view as preliminary due to the limited number of
questions, whereas the Jurisdiction/Time category is excluded
from the analysis because of its very small sample size.
These categories capture distinct legal reasoning demands: FR 
questions primarily require retrieving explicit legal facts, GR 
questions require explaining or synthesizing legal rules and 
conditions, and FP questions require identifying and rejecting 
an incorrect assumption before answering.

For the GR category, Llama3-8B-based systems consistently 
exhibit the highest hallucination rates. E5+Llama3-8B 
performs worst on GDPR (AL = 46.6\%), whereas 
BM25+Llama3-8B has the highest hallucination rate on CIVIL 
(AL = 39.4\%). The relatively large standard deviations of the 
CL scores across both datasets indicate substantial variability 
in claim-level hallucination behavior, suggesting that 
Llama3-8B is particularly prone to hallucinations on GR 
questions regardless of the retriever used.

At the other end of the spectrum, E5+Mixtral-8x7B and 
E5+GPT-4 achieve the lowest AL on GDPR (20.6\%), followed 
closely by BM25+GPT-5 (22.1\%). However, BM25+GPT-5 
attains the lowest CL overall ($1.6{\scriptstyle \pm3.6}$), with 
considerably lower variability than E5+Mixtral-8x7B ($9.4$) 
and E5+GPT-4 ($7.1$). Taken together, these results suggest 
that BM25+GPT-5 is the most robust system for GR 
questions, combining low hallucination frequency with stable 
claim-level behavior, a trend that is further confirmed on CIVIL, 
where it achieves the lowest hallucination rates.

For the FR category, Llama3-8B- and Mixtral-8x7B-based 
systems generally exhibit the highest hallucination rates 
across both GDPR and CIVIL. Notable exceptions are 
BM25+Mixtral-8x7B on GDPR and E5+Mixtral-8x7B on CIVIL, 
both achieving an AL of 0\%. These results suggest that 
retrieval effectiveness may depend on corpus characteristics, 
including differences in language and document structure. In 
contrast, GPT-based generators consistently achieve low 
hallucination rates across both datasets and retrievers, 
indicating greater robustness to variations in retrieval quality. 
Overall, while open-weight models appear more sensitive to 
the choice of retriever and dataset, GPT-based systems 
maintain comparatively stable performance.

We observe for FP category a similar pattern as for FR. 
Llama3-8B and 
Mixtral-8x7B generally exhibit the highest 
hallucination rates across both datasets, with 
BM25+Mixtral-8x7B having AL of $100$\%
on 
CIVIL, i.e., every response to FP questions contained at 
least one hallucinated claim. 
GPT-5 remains 
comparatively 
robust, consistently achieving low hallucination 
rates and reaching $0$\% in several settings.

The FP category appears particularly challenging for 
legal RAG systems as it requires not only 
retrieving relevant legal information but also 
verifying an incorrect assumption. For 
example, the 
question \textit{``How does GDPR require that controllers and 
	processors submit their assessments and safeguards to a 
	public registry for transparency?''} contains a false premise, 
as 
the GDPR imposes no such public registry requirement. 
Many systems 
implicitly accept the false premise and generate answers that 
build 
on it. While GPT-5 
demonstrates  resilience to this failure mode, the different 
results 
across systems call for further research to better understand 
the impact of question types on hallucination.

\begin{table*}[t]
	\centering
	\footnotesize
	\setlength{\tabcolsep}{3pt}
	
	\begin{tabular*}{\textwidth}{@{\extracolsep{\fill}}l
			c@{\hspace{1pt}}c
			c@{\hspace{1pt}}c
			c@{\hspace{1pt}}c
			c@{\hspace{1pt}}c
			c@{\hspace{1pt}}c
			c@{\hspace{1pt}}c}
		\toprule
		
		& \multicolumn{6}{c}{\textbf{GDPR}}
		& \multicolumn{6}{c}{\textbf{CIVIL}} \\
		
		\cmidrule(lr){2-7}
		\cmidrule(lr){8-13}
		
		& \multicolumn{2}{c}{\textbf{LE} ($Q=52$)}
		& \multicolumn{2}{c}{\textbf{CO} ($Q=77$)}
		& \multicolumn{2}{c}{\textbf{CI} ($Q=20$)}
		& \multicolumn{2}{c}{\textbf{LE} ($Q=104$)}
		& \multicolumn{2}{c}{\textbf{CO} ($Q=29$)}
		& \multicolumn{2}{c}{\textbf{CI} ($Q=35$)} \\
		
		\cmidrule(lr){2-3}
		\cmidrule(lr){4-5}
		\cmidrule(lr){6-7}
		\cmidrule(lr){8-9}
		\cmidrule(lr){10-11}
		\cmidrule(lr){12-13}
		
		\textbf{RAG System}
		& CL$\downarrow$ & AL$\downarrow$
		& CL$\downarrow$ & AL$\downarrow$
		& CL$\downarrow$ & AL$\downarrow$
		& CL$\downarrow$ & AL$\downarrow$
		& CL$\downarrow$ & AL$\downarrow$
		& CL$\downarrow$ & AL$\downarrow$ \\
		
		\midrule
		
		BM25+Llama3-8B
		& $4.3{\scriptstyle\pm7.6}$ & $30.8$
		& $4.8{\scriptstyle\pm8.8}$ & $31.2$
		& $4.9{\scriptstyle\pm10.5}$ & $20.0$
		& $6.0{\scriptstyle\pm11.9}$ & $26.2$
		& $36.3{\scriptstyle\pm31.2}$ & $72.4$
		& $37.9{\scriptstyle\pm30.3}$ & $80.0$ \\
		
		BM25+Mixtral-8x7B
		& $4.7{\scriptstyle\pm9.3}$ & $28.9$
		& $3.5{\scriptstyle\pm10.3}$ & $\mathbf{19.5}$
		& $\mathbf{0.0{\scriptstyle\pm0.0}}$ & $\mathbf{0.0}$
		& $5.7{\scriptstyle\pm14.4}$ & $20.4$
		& $28.3{\scriptstyle\pm23.8}$ & $75.9$
		& $37.9{\scriptstyle\pm29.7}$ & $71.4$ \\
		
		BM25+GPT-4
		& $2.5{\scriptstyle\pm3.8}$ & $34.6$
		& $3.3{\scriptstyle\pm5.7}$ & $31.2$
		& $2.2{\scriptstyle\pm4.7}$ & $20.0$
		& $\mathbf{1.7{\scriptstyle\pm6.1}}$ & $\mathbf{10.7}$
		& $13.8{\scriptstyle\pm30.7}$ & $31.0$
		& $3.3{\scriptstyle\pm8.1}$ & $17.1$ \\
		
		BM25+GPT-5
		& $\mathbf{1.6{\scriptstyle\pm3.7}}$ & $23.1$
		& $\mathbf{1.9{\scriptstyle\pm4.3}}$ & $23.4$
		& $\mathbf{0.0{\scriptstyle\pm0.0}}$ & $\mathbf{0.0}$
		& $\mathbf{1.7{\scriptstyle\pm5.8}}$ & $11.7$
		& $\mathbf{0.3{\scriptstyle\pm1.7}}$ & $\mathbf{3.5}$
		& $0.4{\scriptstyle\pm2.1}$ & $2.9$ \\
		
		E5+Llama3-8B
		& $7.9{\scriptstyle\pm10.5}$ & $44.2$
		& $11.6{\scriptstyle\pm13.7}$ & $54.6$
		& $5.8{\scriptstyle\pm10.7}$ & $35.0$
		& $14.7{\scriptstyle\pm25.0}$ & $35.9$
		& $7.4{\scriptstyle\pm12.6}$ & $31.0$
		& $4.3{\scriptstyle\pm8.9}$ & $22.9$ \\
		
		E5+Mixtral-8x7B
		& $1.9{\scriptstyle\pm5.5}$ & $\mathbf{11.5}$
		& $5.9{\scriptstyle\pm10.7}$ & $27.3$
		& $7.5{\scriptstyle\pm14.8}$ & $25.0$
		& $6.2{\scriptstyle\pm16.5}$ & $18.5$
		& $5.5{\scriptstyle\pm19.2}$ & $13.8$
		& $\mathbf{0.0{\scriptstyle\pm0.0}}$ & $\mathbf{0.0}$ \\
		
		E5+GPT-4
		& $2.5{\scriptstyle\pm4.9}$ & $23.1$
		& $3.5{\scriptstyle\pm8.3}$ & $20.8$
		& $2.1{\scriptstyle\pm6.6}$ & $10.0$
		& $3.1{\scriptstyle\pm9.9}$ & $\mathbf{10.7}$
		& $7.2{\scriptstyle\pm21.3}$ & $13.8$
		& $1.0{\scriptstyle\pm5.6}$ & $2.9$ \\
		
		E5+GPT-5
		& $5.3{\scriptstyle\pm8.5}$ & $38.5$
		& $6.7{\scriptstyle\pm9.5}$ & $42.9$
		& $2.5{\scriptstyle\pm11.2}$ & $5.0$
		& $6.0{\scriptstyle\pm11.5}$ & $31.1$
		& $4.3{\scriptstyle\pm19.0}$ & $6.9$
		& $1.9{\scriptstyle\pm8.8}$ & $5.7$ \\
		
		\bottomrule
	\end{tabular*}
	
	\caption{Hallucination rates across \textit{personas }(\%) 
	(\textbf{RQ2}).
	LE = Legal Expert, CO = Civil Officer, and 
	CI = Citizen.}
	\label{tab:hallucination_per_persona}
\end{table*}


\noindent\textbf{Hallucination Analysis by User Persona.}
Table~\ref{tab:hallucination_per_persona} reports CL and AL 
across three personas: 
Legal Expert (LE), Civil Officer (CO), and Citizen (CI). These 
personas capture distinct legal information needs. CI 
questions are primarily fact-oriented and focus on rights, 
obligations, and statutory references, e.g., \emph{``What is 
the minimum age at which a child's consent is considered valid 
for the processing of personal data under the GDPR?''}. CO 
questions emphasize compliance and institutional 
responsibilities, e.g., \emph{``What are the primary objectives 
of the GDPR regarding the protection of natural persons and 
the free movement of personal data?''}. LE questions require 
legal interpretation and the application of legal provisions to 
more complex scenarios, e.g., \emph{``What specific 
conditions must be met for the transfer of personal data to a 
third country without an adequacy decision or appropriate 
safeguards?''}. 

Across both datasets, GPT-based systems generally achieve 
the lowest hallucination rates, with BM25+GPT-5 emerging as 
the most consistent configuration. In contrast, 
Llama3-8B-based systems typically exhibit the highest 
hallucination rates, particularly when combined with E5 
retrieval. 

On GDPR, LE questions typically result in fewer hallucinations 
than CO questions, while some systems achieve their lowest 
rates on CI questions. A similar 
pattern emerges on CIVIL, where differences between 
personas are substantially larger. In particular, CO and CI 
questions prove challenging for several open-weight models. 
For example, BM25+Llama3-8B and BM25+Mixtral-8x7B 
exceed 70\% AL for both personas, whereas GPT-based 
systems remain below 20\% in most cases and often below 
5\% with GPT-5.

Interestingly, LE questions frequently result in lower 
hallucination rates than CO questions despite requiring more 
sophisticated legal reasoning. One possible explanation is that 
expert-authored questions tend to be more precise and legally 
constrained, reducing ambiguity and limiting opportunities for 
unsupported inferences. In contrast, CO questions often 
require reasoning about legal procedures and institutional 
responsibilities, encouraging broader explanations that may 
not be fully grounded in the retrieved evidence. 

Overall, the results suggest that legal RAG systems can 
be sensitive to changes in question formulation and user intent.


\begin{table*}[t]
	\centering
	\footnotesize
	\setlength{\tabcolsep}{4pt}
	
	\begin{tabular*}{\textwidth}{@{\extracolsep{\fill}}lcccccccccc}
		\toprule
		
		& \multicolumn{5}{c}{\textbf{GDPR}}
		& \multicolumn{5}{c}{\textbf{CIVIL}} \\
		
		\cmidrule(lr){2-6}
		\cmidrule(lr){7-11}
		
		\textbf{RAG System}
		& $M$
		& CL$\downarrow$
		& CL$^{s}\downarrow$
		& AL$\downarrow$
		& AL$^{s}\downarrow$
		& $M$
		& CL$\downarrow$
		& CL$^{s}\downarrow$
		& AL$\downarrow$
		& AL$^{s}\downarrow$ \\
		
		\midrule
		
		BM25+Llama3-8B
		& $6.0{\scriptstyle\pm4.8}$
		& $15.8{\scriptstyle\pm26.2}$
		& $7.7{\scriptstyle\pm18.9}$
		& $42.3$
		& $25.4$
		& $4.1{\scriptstyle\pm2.6}$
		& $15.9{\scriptstyle\pm24.8}$
		& $18.0{\scriptstyle\pm27.4}$
		& $40.8$
		& $43.7$ \\
		
		BM25+GPT-4
		& $3.8{\scriptstyle\pm3.4}$
		& $5.8{\scriptstyle\pm15.7}$
		& $9.6{\scriptstyle\pm25.1}$
		& $18.3$
		& $19.7$
		& $2.7{\scriptstyle\pm2.2}$
		& $\mathbf{3.1}{\scriptstyle\pm10.8}$
		& $10.7{\scriptstyle\pm23.7}$
		& $9.9$
		& $23.9$ \\
		
		BM25+GPT-5
		& $5.2{\scriptstyle\pm4.6}$
		& $7.2{\scriptstyle\pm16.5}$
		& $\mathbf{4.5}{\scriptstyle\pm15.3}$
		& $22.5$
		& $\mathbf{12.7}$
		& $4.9{\scriptstyle\pm2.7}$
		& $11.8{\scriptstyle\pm22.2}$
		& $\mathbf{9.0}{\scriptstyle\pm23.7}$
		& $33.8$
		& $\mathbf{19.7}$ \\
		
		E5+Llama3-8B
		& $5.8{\scriptstyle\pm4.9}$
		& $12.6{\scriptstyle\pm24.9}$
		& $7.1{\scriptstyle\pm19.8}$
		& $25.4$
		& $18.3$
		& $4.0{\scriptstyle\pm2.8}$
		& $14.1{\scriptstyle\pm25.1}$
		& $17.5{\scriptstyle\pm32.8}$
		& $33.8$
		& $33.8$ \\
		
		E5+GPT-4
		& $4.4{\scriptstyle\pm4.3}$
		& $\mathbf{1.9}{\scriptstyle\pm10.1}$
		& $9.6{\scriptstyle\pm25.1}$
		& $\mathbf{4.2}$
		& $18.3$
		& $3.0{\scriptstyle\pm1.7}$
		& $\mathbf{3.1}{\scriptstyle\pm11.6}$
		& $15.6{\scriptstyle\pm31.5}$
		& $\mathbf{8.5}$
		& $23.9$ \\
		
		E5+GPT-5
		& $5.3{\scriptstyle\pm5.2}$
		& $5.9{\scriptstyle\pm17.7}$
		& $6.0{\scriptstyle\pm19.7}$
		& $18.3$
		& $\mathbf{12.7}$
		& $4.9{\scriptstyle\pm3.2}$
		& $8.7{\scriptstyle\pm20.7}$
		& $9.1{\scriptstyle\pm21.6}$
		& $25.4$
		& $23.9$ \\
		
		\bottomrule
	\end{tabular*}
	
	\caption{Hallucination rates (\%) of RAG systems evaluated on manually curated legal expert questions within a legal application context. 
		$M$ represents the average number of claims per response, reported as mean $\pm$ standard deviation. 
		}
	\label{tab:external_validation_overall}
	
\end{table*}
\begin{table}[t]
	\centering
	\footnotesize
	\setlength{\tabcolsep}{2pt}
	
	\begin{tabular}{lcccccc}
		\toprule
		
		& \multicolumn{3}{c}{\textbf{GDPR}}
		& \multicolumn{3}{c}{\textbf{CIVIL}} \\
		
		\cmidrule(lr){2-4}
		\cmidrule(lr){5-7}
		
		\textbf{RAG System}
		& \textbf{GR}
		& \textbf{FR} 
		& \textbf{FP} 
		& \textbf{GR}
		& \textbf{FR} 
		& \textbf{FP} \\
		\multicolumn{1}{r}{$Q=$} & $32$ & $12$ & $26$ & $26$ 
		& $7$ & 
		$31$ \\
		\midrule
		
		BM25+Llama3-8B
		& 34.4 & 33.3 & 53.9
		& 30.8 & 71.4 & 41.9 \\
		
		BM25+GPT-4
		& 21.9 & \textbf{0.0} & 23.1
		& 11.5 & \textbf{14.3} & 9.7 \\
		
		BM25+GPT-5
		& 15.6 & 16.7 & 30.8
		& 30.8 & 57.1 & 32.3 \\
		
		E5+Llama3-8B
		& 28.1 & 8.3 & 30.8
		& 19.2 & \textbf{14.3} & 54.8 \\
		
		E5+GPT-4
		& \textbf{0.0} & \textbf{0.0} & \textbf{11.5}
		& \textbf{7.7} & 28.6 & \textbf{6.5} \\
		
		E5+GPT-5
		& 15.6 & 8.3 & 23.1
		& 15.4 & 28.6 & 38.7 \\
		
		\bottomrule
	\end{tabular}
	
	\caption{Answer-level hallucination rates (\%) by question 
	category on expert-authored legal questions.}
	\label{tab:hallucination_per_cat_manual_QA}
	
\end{table}

%
      

\section{Validation on Expert-Authored Questions}
\label{sec:discussion}

To assess the robustness of our findings beyond 
ClaimRAG-LAW~\cite{das2026}, we conducted an external 
validation using 
142 questions manually drafted by a third-party 
legal expert, evenly split between GDPR and CIVIL. 
Table~\ref{tab:external_validation_overall} reports the overall 
hallucination rates, while 
Table~\ref{tab:hallucination_per_cat_manual_QA} provides a 
breakdown by question category. Overall, the principal trends 
observed on ClaimRAG-LAW are reproduced. GPT-based 
systems continue to exhibit the lowest hallucination rates, 
whereas Llama3-8B-based systems are the most prone to 
hallucination. Although the hallucination rates 
differ from those observed on the benchmark, the consistency 
of the system rankings across both legal corpora suggests 
that the observed patterns are not benchmark-specific and 
generalize to independently curated 
legal questions.

The category-level results confirm the difficulty of 
FP questions. Across both GDPR and CIVIL, FP 
questions frequently yield some of the highest answer-level 
hallucination rates, indicating that many systems continue to 
accept incorrect assumptions rather than explicitly rejecting 
them. This finding reinforces the importance of premise 
verification as a key 
challenge for legal RAG systems.

The external validation also provides additional insight into the 
behavior of GPT-4 and GPT-5. While the two models achieve 
broadly comparable overall performance, they exhibit distinct 
hallucination profiles. GPT-4 often attains slightly lower overall 
CL and AL hallucination rates, whereas 
GPT-5 consistently achieves lower strict hallucination rates 
(CL$^s$ and AL$^s$). 
This suggests that GPT-5 is less prone to generating claims 
that directly contradict the available evidence, unlike GPT-4.  
Consequently, 
GPT-5 appears more effective at avoiding severe 
hallucinations, even when the overall hallucination rates of the 
two models are similar.

Taken together, these results indicate that improving retrieval 
quality alone is unlikely to mitigate hallucinations in legal RAG 
systems. Future research should therefore focus not only on 
retrieval effectiveness, but also on improving faithfulness to 
retrieved evidence and strengthening premise-verification 
capabilities, particularly for questions containing misleading or 
incorrect assumptions.

			
			\section{Threats to Validity and Limitations}
\label{sec:threats}

\textit{Internal Validity. }
Our analysis relies on the benchmark data and 
diagnostic metrics reported in 
ClaimRAG-LAW~\cite{das2026}. Consequently, 
errors in the benchmark annotations or reported 
metrics may affect the hallucination rates and analyses presented in 
this study. To 
mitigate this threat, we relied on the benchmark's 
expert-validated QA pairs and standardized 
evaluation pipeline, which ensure a consistent 
comparison across all analyzed RAG systems. 

\noindent \textit{External Validity.} Our study is limited 
to two 
legal sources (GDPR and CIVIL) and eight RAG 
systems. Consequently, the findings may not 
generalize to other jurisdictions, languages, retrievers, 
or foundation models. We partially mitigated this threat 
by evaluating systems across datasets from different 
jurisdictions and languages, as well as multiple 
retriever-generator combinations. 

\noindent \textit{Limitations.} Our analysis is based on 
the 
benchmark outputs and therefore does not assess the 
impact of alternative prompts, chunking strategies, 
retrieval parameters, or newer model versions. 
Furthermore, some question categories, particularly 
False Premise and Jurisdiction/Time-Specific 
questions, contain relatively few examples, which may 
affect the stability of the corresponding results. Future 
work should extend the evaluation to additional legal 
corpora, larger datasets, and expert-reviewed 
hallucination annotations.


		\section{Conclusion}
\label{sec:conclusion}

We have presented a fine-grained analysis of hallucinations in eight 
legal retrieval-augmented generation (RAG) systems 
evaluated on two legal corpora, GDPR and CIVIL. Building on 
ClaimRAG-LAW~\cite{das2026}, our study has characterized 
hallucination behavior through complementary answer-level 
and claim-level analyses.

Our results show that hallucinations remain a persistent 
challenge in legal RAG systems. Hallucination rates vary 
substantially across configurations, ranging from the most 
robust system, BM25+GPT-5, to the least robust, 
Llama3-8B-based systems, which can hallucinate in nearly 
half of all generated 
responses. We further find that most hallucinations consist of 
isolated unsupported claims rather than pervasive errors 
throughout an answer. Moreover, hallucination behavior varies 
substantially across question categories and user personas, 
with factual-recall and false-premise questions, as well as 
queries posed by non-expert users, proving particularly 
challenging.
By validating our findings on an independent set of 
expert-authored legal questions, we provide evidence that the 
observed trends are not specific to ClaimRAG-LAW and 
generalize beyond the benchmark.

In future work, we plan to extend this analysis with 
legal-specific evaluation metrics and additional diagnostic 
studies to better understand the factors contributing to 
hallucinations in legal RAG systems and to support the 
development of more reliable AI-assisted legal applications.


			\bibliographystyle{plainnat}	
			\bibliography{main}

@misc{hu2024,
 title={RefChecker: Reference-based Fine-grained Hallucination 
 Checker and Benchmark for Large Language Models}, 
 author={Xiangkun Hu and Dongyu Ru and Lin Qiu and Qipeng Guo 
 and Tianhang Zhang and Yang Xu and Yun Luo and Pengfei Liu and 
 Yue Zhang and Zheng Zhang},
 year={2024},
 eprint={2405.14486},
 archivePrefix={arXiv},
 primaryClass={cs.CL},
 url={https://arxiv.org/abs/2405.14486}, 
 }

@article{ru2024,
  title={{RAGChecker}: A fine-grained framework for 
  diagnosing 
  retrieval-augmented generation},
  author={Ru, Dongyu and Qiu, Lin and Hu, Xiangkun and Zhang, Tianhang and Shi, Peng and Chang, Shuaichen and Jiayang, Cheng and Wang, Cunxiang and Sun, Shichao and Li, Huanyu and others},
  journal={Advances in Neural Information Processing Systems},
  volume={37},
  pages={21999--22027},
  year={2024},
  url={https://doi.org/10.52202/079017-0692}
}

@article{magesh2025,
  title={Hallucination-Free? Assessing the Reliability of Leading 
  {AI} Legal Research Tools},
  author={Magesh, Varun and Surani, Faiz and Dahl, Matthew and Suzgun, Mirac and Manning, Christopher D and Ho, Daniel E},
  journal={Journal of Empirical Legal Studies},
  volume={22},
  number={2},
  pages={216--242},
  year={2025},
  publisher={Wiley Online Library},
  url={https://doi.org/10.1111/jels.12413}
}

@article{dahl2024,
	title={Large legal fictions: Profiling legal hallucinations in large language models},
	author={Dahl, Matthew and Magesh, Varun and Suzgun, Mirac and Ho, Daniel E},
	journal={Journal of Legal Analysis},
	volume={16},
	number={1},
	pages={64--93},
	year={2024},
	publisher={Oxford University Press UK},
  url={https://doi.org/10.1093/jla/laae003}
}

@misc{pipitone2024,
      title={LegalBench-RAG: A Benchmark for Retrieval-Augmented 
      Generation in the Legal Domain}, 
author={Nicholas Pipitone and Ghita Houir Alami},
year={2024},
eprint={2408.10343},
archivePrefix={arXiv},
primaryClass={cs.AI},
url={https://arxiv.org/abs/2408.10343}, 
}

@inproceedings{niklaus2023,
	title={{LEXTREME}: A Multi-Lingual and Multi-Task Benchmark for the Legal Domain},
	author={Niklaus, Joel and Matoshi, Veton and Rani, Pooja 
	and Galassi, Andrea and St{\"u}rmer, Matthias and Chalkidis, 
	Ilias},
	booktitle={Findings of the Association for Computational Linguistics: EMNLP 2023},
	pages={3016--3054},
	year={2023},
  url={https://doi.org/10.18653/v1/2023.findings-emnlp.200}
}

@inproceedings{chalkidis2022,
  title={{LexGLUE}: A benchmark dataset for legal language 
  understanding in English},
  author={Chalkidis, Ilias and Jana, Abhik and Hartung, Dirk and Bommarito, Michael and Androutsopoulos, Ion and Katz, Daniel and Aletras, Nikolaos},
  booktitle={Proceedings of the 60th Annual Meeting of the Association for Computational Linguistics (Volume 1: Long Papers)},
  pages={4310--4330},
  year={2022},
  url={https://doi.org/10.18653/v1/2022.acl-long.297}
}

@inproceedings{louis2024,
  title={Interpretable long-form legal question answering with retrieval-augmented large language models},
  author={Louis, Antoine and van Dijck, Gijs and Spanakis, Gerasimos},
  booktitle={Proceedings of the AAAI Conference on Artificial Intelligence},
  xxvolume={38},
  xxnumber={20},
  pages={22266--22275},
  year={2024},
  url={https://doi.org/10.1609/aaai.v38i20.30232}
}

@inproceedings{min2023,
  title={{FActScore}: Fine-grained atomic evaluation of factual 
  precision in long form text generation},
  author={Min, Sewon and Krishna, Kalpesh and Lyu, Xinxi and Lewis, Mike and Yih, Wen-tau and Koh, Pang and Iyyer, Mohit and Zettlemoyer, Luke and Hajishirzi, Hannaneh},
  booktitle={Proceedings of the 2023 Conference on Empirical 
  Methods in Natural Language Processing},
  pages={12076--12100},
  year={2023},
  url={https://doi.org/10.18653/v1/2023.emnlp-main.741}
}

@inproceedings{scire2024,
  title={{FENICE}: Factuality Evaluation of summarization based 
  on Natural language Inference and Claim Extraction},
  author={Scir{\`e}, Alessandro and Ghonim, Karim and Navigli, Roberto},
  booktitle={Findings of the Association for Computational Linguistics ACL 2024},
  pages={14148--14161},
  year={2024},
  url={https://doi.org/10.18653/v1/2024.findings-acl.841}
}

@article{robertson2009,
author = {Robertson, Stephen and Zaragoza, Hugo},
title = {The Probabilistic Relevance Framework: {BM25} and 
Beyond},
year = {2009},
issue_date = {April 2009},
publisher = {Now Publishers Inc.},
address = {Hanover, MA, USA},
volume = {3},
number = {4},
issn = {1554-0669},
url = {https://doi.org/10.1561/1500000019},
xxdoi = {10.1561/1500000019},
journal = {Found. Trends Inf. Retr.},
month = apr,
pages = {333–389},
numpages = {57}
}

@inproceedings{wang2024,
	title = "Improving Text Embeddings with Large Language Models",
	author = "Wang, Liang  and
	Yang, Nan  and
	Huang, Xiaolong  and
	Yang, Linjun  and
	Majumder, Rangan  and
	Wei, Furu",
	editor = "Ku, Lun-Wei  and
	Martins, Andre  and
	Srikumar, Vivek",
	booktitle = "Proceedings of the 62nd Annual Meeting of the 
	Association for Computational Linguistics (Volume 1: Long 
	Papers)",
	month = aug,
	year = "2024",
	address = "Bangkok, Thailand",
	publisher = "Association for Computational Linguistics",
	url = "https://aclanthology.org/2024.acl-long.642/",
	doi = "10.18653/v1/2024.acl-long.642",
	pages = "11897--11916",
}

@misc{jiang2024,
 title={Mixtral of Experts}, 
 author={Albert Q. Jiang and others},
 year={2024},
 eprint={2401.04088},
 archivePrefix={arXiv},
 primaryClass={cs.LG},
 url={https://arxiv.org/abs/2401.04088}, 
 }

@misc{grattafiori2024,
 title={The Llama 3 Herd of Models},
 author={Aaron Grattafiori and others},
 year={2024},
 eprint={2407.21783},
 archivePrefix={arXiv},
 primaryClass={cs.AI},
 url={https://arxiv.org/abs/2407.21783}, 
 }

@misc{gpt4,
	title={GPT-4 Technical Report}, 
	author={OpenAI and others},
	year={2024},
	eprint={2303.08774},
	archivePrefix={arXiv},
	primaryClass={cs.CL},
	url={https://arxiv.org/abs/2303.08774}, 
}

@misc{EUGDPR,
	author = {{The European Parliament and the Council of the 
	European Union}},
	title = {Regulation (EU) 2016/679 of the European Parliament 
	and of the Council of 27 April 2016 on the protection of 
	natural persons with regard to the processing of personal 
	data and on the free movement of such data, and repealing 
	Directive 95/46/EC 
	({General Data Protection Regulation})},
	number = {119},
	year = {2016},
	month = {05},
	pages = {1-88},
	url = {https://eur-lex.europa.eu/eli/reg/2016/679/oj}
}

@misc{CIVIL,
	title     = {National Civil Code (consolidated version)},
	author    = {{Grand-Duch{\'{e}} de Luxembourg}},
	publisher = {Journal officiel du Grand-Duch{\'{e}} de 
	Luxembourg (Legilux)},
	url       = 
	{https://legilux.public.lu/},
	urldate   = {2026-01-19},
	year = {2025},
}

@inproceedings{manakul2023,
  title={{SelfCheckGPT}: Zero-resource black-box hallucination 
  detection for generative large language models},
  author={Manakul, Potsawee and Liusie, Adian and Gales, Mark},
  booktitle={Proceedings of the 2023 conference on empirical methods in natural language processing},
  pages={9004--9017},
  year={2023},
  url={https://doi.org/10.18653/v1/2023.emnlp-main.557}
}

@misc{DATASET,
	title = {{ClaimRAG-LAW Dataset}},
	author = {Das, Souvick and Abualhaija, Sallam and Bianculli, 
	Domenico},
	howpublished = {\url{LINK}},
	year         = {2026},
	doi          = {LINK},
}

@misc{singh2025,
  title={OpenAI GPT-5 System Card}, 
  author={Aaditya Singh and others},
  year={2026},
  eprint={2601.03267},
  archivePrefix={arXiv},
  primaryClass={cs.CL},
  url={https://arxiv.org/abs/2601.03267}, 
  }

@inproceedings{hu2025,
  title={Fine-tuning large language models for improving factuality in legal question answering},
  author={Hu, Yinghao and Gan, Leilei and Xiao, Wenyi and Kuang, Kun and Wu, Fei},
  booktitle={Proceedings of the 31st international conference on computational linguistics},
  pages={4410--4427},
  year={2025}
}

@inproceedings{galimzianova2025,
  title={From RAG to Reality: Coarse-Grained Hallucination Detection via NLI Fine-Tuning},
  author={Galimzianova, Daria and Boriskin, Aleksandr and Arshinov, Grigory},
  booktitle={Proceedings of the Fifth Workshop on Scholarly Document Processing (SDP 2025)},
  pages={353--359},
  year={2025},
  url={https://doi.org/10.18653/v1/2025.sdp-1.34}
}

@article{huang2025,
  title={A survey on hallucination in large language models: Principles, taxonomy, challenges, and open questions},
  author={Huang, Lei and Yu, Weijiang and Ma, Weitao and Zhong, Weihong and Feng, Zhangyin and Wang, Haotian and Chen, Qianglong and Peng, Weihua and Feng, Xiaocheng and Qin, Bing and others},
  journal={ACM Transactions on Information Systems},
  volume={43},
  number={2},
  pages={1--55},
  year={2025},
  url={https://doi.org/10.1145/3703155},
  publisher={ACM New York, NY}
}

@inproceedings{enguehard2025,
  title={LeMAJ (Legal LLM-as-a-Judge): Bridging Legal Reasoning and LLM Evaluation},
  author={Enguehard, Joseph and Van Ermengem, Morgane and Atkinson, Kate and Cha, Sujeong and Chowdhury, Arijit Ghosh and Ramaswamy, Prashanth Kallur and Roghair, Jeremy and Marlowe, Hannah R and Negreanu, Carina Suzana and Boxall, Kitty and others},
  booktitle={Proceedings of the Natural Legal Language Processing Workshop 2025},
  pages={318--337},
  year={2025},
  url={https://doi.org/10.18653/v1/2025.nllp-1.23}
}

@misc{das2026,
title={Fine-grained Claim-level RAG Benchmark for Law}, 
author={Souvick Das and Sallam Abualhaija and Domenico Bianculli},
year={2026},
eprint={2605.21071},
archivePrefix={arXiv},
primaryClass={cs.CL},
url={https://arxiv.org/abs/2605.21071}, 
}

@inproceedings{yeh2026,
	title={{LUMINA}: Detecting Hallucinations in {RAG} System with 
	Context{\textendash}Knowledge Signals},
	author={Samuel Yeh and Sharon Li and Tanwi Mallick},
	booktitle={The Fourteenth International Conference on Learning 
	Representations},
	year={2026},
	url={https://openreview.net/forum?id=oJgNNBNEJM}
}

@inproceedings{hu2026,
  title={Detecting hallucinations in retrieval-augmented generation via semantic-level internal reasoning graph},
  author={Hu, Jianpeng and Li, Yanzeng and Zhong, Jialun and Zou, Lei and Qi, Wenfa},
  booktitle={Findings of the Association for Computational Linguistics: ACL 2026},
  pages={27826--27841},
  year={2026}
}

@misc{liu2026,
title={Who Checks the Citations? Benchmarking Legal Hallucination 
Detection}, 
author={Patty Liu and Dominik Stammbach and Peter Henderson},
year={2026},
eprint={2606.21155},
archivePrefix={arXiv},
primaryClass={cs.CL},
url={https://arxiv.org/abs/2606.21155}, 
}

@inproceedings{Li2024,
	title     = {Reference-free Hallucination Detection for Large 
	Vision-Language Models},
	author    = {Li, Qing and Geng, Jiahui and Lyu, Chenyang and 
	Zhu, Derui and Panov, Maxim and Karray, Fakhri},
	booktitle = {Proceedings of the 2024 Conference on 
	Empirical Methods in Natural Language Processing: 
	Findings},
	year      = {2024},
	publisher = {Association for Computational Linguistics},
	pages      = {4542--4551},
	doi        = {10.18653/v1/2024.findings-emnlp.262},
	url        = {https://aclanthology.org/2024.findings-emnlp.262}
}

@inproceedings{orgad2025,
	title={{LLMs} know more than they show: On the intrinsic 
	representation of {LLM} hallucinations},
	author={Orgad, Hadas and Toker, Michael and Gekhman, 
	Zorik and Reichart, Roi and Szpektor, Idan and Kotek, Hadas 
	and Belinkov, Yonatan},
	booktitle={International Conference on Learning 
	Representations},
	volume={2025},
	pages={66880--66913},
	year={2025}
}

@inproceedings{bang2025,
	title={Hallulens: Llm hallucination benchmark},
	author={Bang, Yejin and Ji, Ziwei and Schelten, Alan and 
	Hartshorn, Anthony and Fowler, Tara and Zhang, Cheng and 
	Cancedda, Nicola and Fung, Pascale},
	booktitle={Proceedings of the 63rd Annual Meeting of the 
	Association for Computational Linguistics (Volume 1: Long 
	Papers)},
	pages={24128--24156},
	year={2025}
}

@InProceedings{banerjee2025,
	author="Banerjee, Sourav
	and Agarwal, Ayushi
	and Singla, Saloni",
	title="LLMs Will Always Hallucinate, and We Need to Live 
	with This",
	booktitle="Intelligent Systems and Applications",
	year="2025",
	publisher="Springer Nature Switzerland",
	pages="624--648",
}

@misc{gcai2026,
	author = {{GC AI}},
	title = {AI Hallucinations in Legal Cases: A Sanctions Tracker 
	(2026)},
	year = {2026},
	howpublished = 
	{\url{https://gc.ai/blog/ai-hallucination-legal-cases}},
	note = {Accessed: 2026-07-22}
}

@inproceedings{mik2024,
	author = {Mik, Eliza},
	title = {The Ground Truth about Legal Hallucinations},
	booktitle = {ICML 2024 Workshop on Generative AI and Law 
	(GenLaw)},
	year = {2024},
	month = jul
}

@article{latif2025,
	author = {Latif, Youssef Abdel},
	title   = {Hallucinations in Large Language Models and Their 
	Influence on Legal Reasoning: Examining the Risks of 
	AI-Generated Factual Inaccuracies in Judicial Processes},
	journal = {Journal of Computational Intelligence, Machine 
	Reasoning, and Decision-Making},
	volume  = {10},
	number  = {2},
	pages   = {10--20},
	year    = {2025},
	month   = feb,
	url     = 
	{https://morphpublishing.com/index.php/JCIMRD/article/view/2025-02-07}
}

@inproceedings{hou2024,
	title = "Gaps or Hallucinations? Scrutinizing 
	Machine-Generated Legal Analysis for Fine-grained Text 
	Evaluations",
	author = "Hou, Abe  and
	Jurayj, William  and
	Holzenberger, Nils  and
	Blair-Stanek, Andrew  and
	Van Durme, Benjamin",
	booktitle = "Proceedings of the Natural Legal Language 
	Processing Workshop 2024",
	month = nov,
	year = "2024",
	address = "Miami, FL, USA",
	publisher = "Association for Computational Linguistics",
	url = "https://aclanthology.org/2024.nllp-1.24/",
	doi = "10.18653/v1/2024.nllp-1.24",
	pages = "280--302"
}

@inproceedings{lin2025,
	title = "Persona-{SQ}: A Personalized Suggested Question 
	Generation Framework For Real-world Documents",
	author = "Lin, Zihao  and
	Wang, Zichao  and
	Pan, Yuanting  and
	Manjunatha, Varun  and
	Rossi, Ryan A.  and
	Lau, Angela  and
	Huang, Lifu  and
	Sun, Tong",
	booktitle = "Proceedings of the 2025 Conference of the 
	Nations of the Americas Chapter of the Association for 
	Computational Linguistics: Human Language Technologies 
	(System Demonstrations)",
	month = apr,
	year = "2025",
	publisher = "Association for Computational Linguistics",
	url = "https://aclanthology.org/2025.naacl-demo.20/",
	doi = "10.18653/v1/2025.naacl-demo.20",
	pages = "210--247",

}

		\end{document}